\documentclass[sigconf]{acmart}
\AtBeginDocument{%
  }
\usepackage{booktabs}
\usepackage{multirow}
\usepackage{xcolor}
\usepackage[colorinlistoftodos]{todonotes}
\usepackage{caption}

\newcommand{\guy}[1]{}
\newcommand{\bracha}[1]{}
\newcommand{\neomi}[1]{}

\begin{document}

\title{EncodeRec: An Embedding Backbone for Recommendation Systems}

\author{Guy Hadad}
\email{guyhada@post.bgu.ac.il}
\affiliation{%
  \institution{Ben-Gurion University of the Negev}
  \city{Be'er Sheva}
  \country{Israel}
}

\author{Neomi Rabaev}
\email{rabaevn@post.bgu.ac.il}
\affiliation{%
  \institution{Ben-Gurion University of the Negev}
  \city{Be'er Sheva}
  \country{Israel}
}

\author{Bracha Shapira}
\email{bshapira@bgu.ac.il}
\affiliation{%
  \institution{Ben-Gurion University of the Negev}
  \city{Be'er Sheva}
  \country{Israel}
}

\renewcommand{\shortauthors}{Hadad et al.}

\begin{abstract}
Recent recommender systems increasingly leverage embeddings from large pre-trained language models (PLMs). However, such embeddings exhibit two key limitations: (1) PLMs are not explicitly optimized to produce structured and discriminative embedding spaces, and (2) their representations remain overly generic, often failing to capture the domain-specific semantics crucial for recommendation tasks.
We present \textbf{EncodeRec}, an approach designed to align textual representations with recommendation objectives while learning compact, informative embeddings directly from item descriptions. EncodeRec keeps the language model parameters frozen during recommender system training, making it computationally efficient without sacrificing semantic fidelity. Experiments across core recommendation benchmarks demonstrate its effectiveness both as a backbone for sequential recommendation models and for semantic ID tokenization, showing substantial gains over PLM-based and embedding model baselines. These results underscore the pivotal role of embedding adaptation in bridging the gap between general-purpose language models and practical recommender systems.
\end{abstract}





\maketitle

\section{Introduction}

Text has become a first-class signal in modern recommender systems. Item catalogs provide rich textual descriptions, and users express preferences through reviews, queries, and conversations. The rapid progress in natural language processing (NLP), particularly the rise of large pre-trained language models (PLMs), has revitalized content-based recommendation \cite{musto2012semantics}: systems can now represent and reason over text with far greater semantic fidelity, and PLM embeddings are widely adopted as off-the-shelf representations for candidate retrieval, reranking, and generative recommenders. However, this shift also exposes a growing mismatch between how PLMs are trained and what recommendation ultimately requires.

We argue that two issues lie at the core of this mismatch. \textbf{First}, while models such as BERT and its variants capture rich semantic structure, PLMs are not trained with metric-learning or retrieval objectives and hence do not produce calibrated, discriminative embedding spaces: their hidden representations are optimized for language modeling rather than for well-separated embeddings tailored to retrieval or ranking tasks. \textbf{Second}, even when repurposed as embeddings, PLM representations remain \emph{generic}: they capture broad topical similarity but underrepresent the fine-grained, domain-specific attributes that drive recommendation quality, such as cross-item substitutability and complementarity, and nuanced preference signals that correlate with user–item interaction patterns.

For example, a PLM may embed a 50-piece and a 1,000-piece puzzle very closely because both belong to the same category. A recommender, however, must distinguish them along dimensions such as difficulty, target age group, price range, or intended purpose attributes that directly affect user decisions. Capturing these fine-grained semantics is essential for accurately modeling user preferences. To serve as an effective backbone for recommendation, embeddings must therefore move beyond surface similarity and encode the underlying attributes often expressed in item titles, descriptions, and structured metadata that truly differentiate items in user decision contexts.

Previous work has frequently adopted embeddings from pre-trained language models (e.g., BERT \cite{devlin2019bert}) without any adaptation, using them as the backbone for recommendation systems \cite{hou2022towards, hou2023learning}. BLaIR \cite{hou2024bridging} takes a step further by adapting PLM-based embeddings through correlations between item metadata and user reviews. However, recent evidence \cite{attimonelli2025we} shows that PLM-based representations still lag behind modern general-purpose embedding models on text retrieval benchmarks, suggesting that existing adaptation strategies are not sufficient. This gap is illustrated in Table~\ref{tab:beauty_retrieval}, which compares BERT, BLaIR, and two general embedding models of different sizes \cite{vera2025embeddinggemma} on a retrieval task where the item title serves as the query and the item description as the target document, illustrating how much room there is to re-think the embedding backbone used in recommendation.

\begin{table}[t]
\centering
\setlength{\tabcolsep}{4pt}
\renewcommand{\arraystretch}{1.1}
\begin{tabular}{lcc}
\toprule
\textbf{Model} & \textbf{Recall@10} & \textbf{NDCG@10} \\
\midrule
BERT-Base & 0.09 & 0.07 \\
BLaIR-Large & 0.12 & 0.09 \\
all-MiniLM-L6-v2 & 0.57 & 0.44 \\
EmbeddingGemma & 0.83 & 0.72 \\
\bottomrule
\end{tabular}
\caption{Retrieval results for item descriptions retrieved from titles on 50K items in the Amazon \textit{Beauty} dataset.}
\label{tab:beauty_retrieval}
\end{table}

In this paper, we introduce EncodeRec, an embedding backbone and training approach specifically designed to produce recommendation-aligned item representations. EncodeRec employs a shared encoder trained with a contrastive objective over item metadata, such as using item titles as queries and their descriptions as corresponding documents, to structure the embedding space around attributes that matter for modeling user preferences. Ultimately, this yields finer-grained item representations, a property that is crucial for a wide range of recommendation tasks. We demonstrate how these embeddings can serve as a drop-in backbone for sequential recommenders such as UniSRec~\cite{hou2022towards}, where they replace item embeddings, and for generative recommendation models such as TIGER \cite{rajput2023recommender}, where they provide the basis for the semantic ID tokenization process, so they improve performance and can eliminate Semantic ID collisions entirely. Across multiple domains, we evaluate EncodeRec against the original backbone model BLaIR and strong general-purpose text embedding models, and show consistent improvements in downstream recommendation quality.

\section{Related Works}

\subsection{Text Embedding Models}

In recent years, the field of text embeddings has advanced substantially. A primary driver of this progress has been the rise of pre-trained language models, which can effectively encode the semantic structure of text into vector representations \cite{reimers2019sentence, ni2022large, karpukhin2020dense}. Another critical factor has been the success of contrastive learning, particularly improvements in negative sampling strategies \cite{xiongapproximate, qu2021rocketqa} and the use of knowledge distillation to transfer representation quality across models \cite{hofstatter2021efficiently, ren2021rocketqav2, zhang2021adversarial}. Building on these foundations, there is a growing emphasis on developing versatile embedding models that can serve a broad range of downstream applications. Notable recent contributions include Contriever \cite{izacard2021unsupervised} and E5 \cite{wang2022text}, as well as more recent large-scale efforts such as Qwen3-Embeddings \cite{zhang2025qwen3}, and EmbeddingGemma \cite{vera2025embeddinggemma}, which significantly extend the coverage and quality of general-purpose text embeddings.

Despite these advances, most general-purpose embedding models are trained on broad web corpora and are not explicitly optimized to capture the nuanced semantics that are critical for recommendation tasks, such as fine-grained item attributes, domain-specific terminology, and subtle signals that shape user–item relevance.

\subsection{Text Encoders for Recommendation}

Text encoders for recommendation typically follow two lines of work: (i) \emph{frozen} language-model (LM) embeddings used as-is, e.g., UniSRec~\cite{hou2022towards} and VQ-Rec~\cite{hou2023learning}; and (ii) \emph{task-tuned} encoders that adapt the LM to the recommendation setting, such as Recformer~\cite{li2023text} and X-Cross~\cite{hadad2025x}. Frozen encoders are attractive for large-scale systems because they are easy to deploy, cost-efficient, and stable at inference time, but they may not fully capture domain-specific semantics needed for high-quality recommendation.

Many prior methods adopt BERT as the textual backbone, but BERT's masked-language-model pretraining does not directly optimize for embedding quality under semantic similarity, which can hinder retrieval performance~\cite{reimers2019sentence, gao2021simcse}. To mitigate this mismatch, BLaIR~\cite{hou2024bridging} trains a contrastive encoder using user reviews as pseudo-queries and item descriptions as documents, improving retrieval in recommendation and search contexts. Conversely, recent studies find that strong general-purpose text embeddings can be competitive without domain-specific specialization~\cite{attimonelli2025we}, especially when downstream models are powerful enough to compensate for representation gaps.

Our work explores a complementary direction: instead of learning query–item alignment from user reviews, we learn item representations grounded purely in objective metadata. While models such as \textit{BLaIR} align item descriptions with user reviews, this coupling introduces subjective, preference-driven noise and can limit generalization. By contrast, we focus on stable textual signals that capture an item's factual identity.

\section{Methodology}

Our goal is to learn item representations that encode the item’s intrinsic semantic content rather than user-dependent linguistic patterns. To this end, we treat the product title as a concise, high-precision semantic anchor, and align it with the item description and structured attributes using a metadata-driven contrastive objective. This formulation removes dependence on user language and encourages the encoder to model the underlying product identity. Because our approach relies solely on objective metadata, the resulting representations are more consistent, less sensitive to review noise, and more reusable as a universal backbone across diverse recommendation tasks.

\begin{figure}[t]
    \centering
    \includegraphics[width=0.9\linewidth]{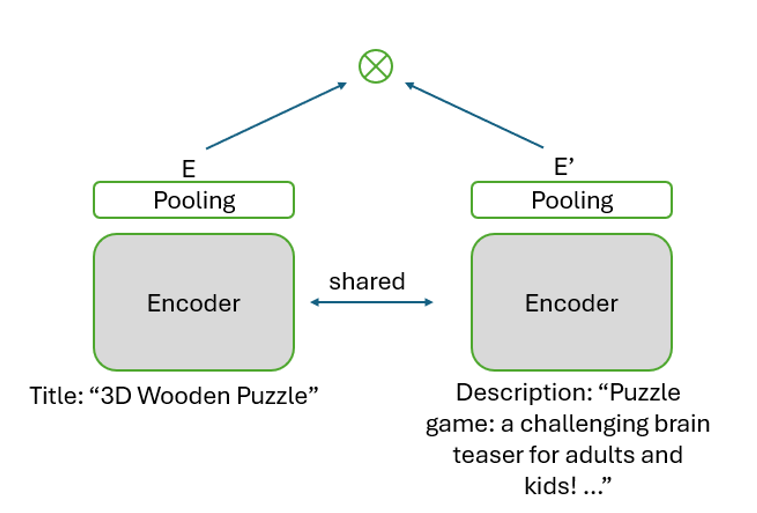}
    \caption{Our metadata-contrastive training approach: concise semantic anchors (titles) are aligned with richer textual descriptions and attributes. Other items in the batch serve as negatives under the InfoNCE objective.}
    \label{fig:method}
\end{figure}

We adopt a single pretrained encoder to embed both metadata views. Using shared parameters ensures both the title and description lie in the same representational space while keeping training lightweight. For a batch of $N$ items, each with title $x_i^{(q)}$ and description+attributes $x_i^{(d)}$, we compute:
\[
h_i^{(q)} = \mathrm{Emb}(x_i^{(q)}), \qquad 
h_i^{(d)} = \mathrm{Emb}(x_i^{(d)}).
\]
Both embeddings are processed through a pooling layer and $L_2$-normalized. We then compute their cosine similarity, which serves as the similarity score within the contrastive learning objective. The full contrastive setup is illustrated in Figure~\ref{fig:method}.

Training is performed with an \textit{in-batch} InfoNCE objective~\cite{rusak2025infonce}, where each title is paired with its corresponding description as the positive example, and all other descriptions in the batch act as negatives:
\[
\mathcal{L} = - \frac{1}{N} \sum_{i=1}^{N} 
\log 
\frac{
\exp\!\left( \mathrm{sim}(h_i^{(q)}, h_i^{(d)}) / \tau \right)
}{
\sum_{j=1}^{N} \exp\!\left( \mathrm{sim}(h_i^{(q)}, h_j^{(d)}) / \tau \right)
},
\]
with temperature $\tau$ controlling the sharpness of the similarity distribution. This objective explicitly consolidates metadata views of the same product while separating unrelated items, yielding a structured and discriminative embedding space tailored for recommendation systems.

\section{Experiments and Discussion}
\subsection{Experimental Setup}

We fine-tune two embedding models of different scales, MiniLM-L6-v2 (22M) and EmbeddingGemma (300M) \cite{vera2025embeddinggemma}. Each model is trained over the Amazon 2023 sub-dataset \cite{hou2024bridging}. For every item, we extract its \emph{title}, \emph{features}, and \emph{description}, concatenating the latter two to form the document text. Training uses in-batch negatives with a batch size of 512, learning rate $5\times10^{-5}$, a warmup ratio of 0.1, ensuring that any performance gains can be attributed solely to the embedding backbone.\footnote{\url{https://huggingface.co/collections/guyhadad01/encodrec}}

To isolate the effect of embeddings, we adopt experimental settings consistent with prior work. For sequential recommendation, we follow the configuration of \textit{UniSRec} \cite{hou2022unisrec} and the setup from \cite{hou2024bridging}. For semantic ID tokenization, we use \textit{TIGER} \cite{rajput2023recommender} with the implementation from \cite{yang2024unifying}.

For baselines, we use the encoders employed in the original works, BERT \cite{devlin2019bert} for UniSRec and Sentence-T5 \cite{ni2022sentence} for TIGER, together with BLAIR-Base and BLAIR-Large \cite{hou2024bridging} as strong embedding-focused baselines. We additionally evaluate our contrastively tuned embeddings (EncodeRec), which adapt general-purpose encoders to item metadata.

Following prior work, TIGER experiments use the \textbf{Sports}, \textbf{Toys}, and \textbf{Beauty} domains from Amazon-2018 \cite{ni2019justifying}, while UniSRec experiments use \textbf{Baby Products}, \textbf{Video Games}, and \textbf{Beauty} from Amazon-2023 \cite{hou2024bridging}. Evaluation follows standard protocols and reports \textbf{Recall@10} and \textbf{nDCG@10} to enable direct and consistent comparison across models.

\subsection{Results and Analysis}
\textbf{Item Embeddings for Sequential Recommendation.}
From Table~\ref{tab:unsirec}, we observe that EncodeRec achieves the highest performance across all datasets, outperforming the other baselines by \textbf{5--26\%}. Notably, the \textit{EncodeRec-22M} model obtains competitive results compared to \textit{BLAIR}, despite using \textbf{6$\times$ to 16$\times$ fewer parameters}. This demonstrates the efficiency of our embedding backbone.

EncodeRec further shows strong \textbf{scalability}, both in terms of model capacity and the amount of domain data available. Increasing the backbone size from 22M to 300M consistently improves performance, a trend that is not observed in BLAIR. Moreover, in the \textit{Beauty} dataset, which contains substantially more training data, the relative improvement of the 22M model over BLAIR is even larger, suggesting that EncodeRec makes better use of additional domain data and is inherently scalable.

To further examine scalability, we also evaluated a larger \textit{Qwen3-0.6B} \cite{zhang2025qwen3} on the \textit{Beauty} dataset and observed the same performance trend. Due to limited computational resources, we did not extend these larger-scale experiments across all datasets, but the results on Beauty provide additional evidence that the scalability of EncodeRec holds for larger backbone models as well.

\textbf{Semantic ID Tokenization.} We adopt \textit{TIGER}, a state-of-the-art generative recommendation framework. In TIGER, items are first encoded into dense embeddings using a pre-trained text encoder and then quantized via an RQ-VAE \cite{lee2022autoregressive} to produce discrete Semantic IDs. These IDs capture hierarchical semantics, coarse at early tokens and increasingly fine-grained at later ones, and are generated autoregressively by a Transformer to predict the next item. Item embeddings therefore form the backbone of the entire pipeline: they are quantized into tokens, sequenced for training, and decoded back into items at inference time. Since the Semantic IDs are derived directly from the embeddings, TIGER is highly sensitive to the quality of the underlying representation.

As shown in Table~\ref{tab:tiger}, using EncodeRec as the embedding backbone yields consistent improvements of \textbf{4--9\%} over the strongest baselines, following the same scalability trend observed in our UniSRec experiments. We additionally evaluated a larger \textit{Qwen3-0.6B} on the \textit{Beauty} dataset and observed the same pattern, further supporting the robustness of this trend.

Crucially, EncodeRec eliminates Semantic ID collisions entirely. This addresses one of the most core weaknesses of existing tokenization-based generative recommenders, where multiple items are often mapped to the same Semantic ID due to limitations in the embedding space. Removing these collisions avoids a major source of ambiguity and error in Semantic ID tokenization.

\begin{table}[t]
\centering
\renewcommand{\arraystretch}{1.6}
\resizebox{\columnwidth}{!}{%
\begin{tabular}{lcccccc}
\toprule
\multirow{2}{*}{\textbf{UniSRec Model}} &
\multicolumn{2}{c}{\textbf{Video Games}} &
\multicolumn{2}{c}{\textbf{Beauty}} &
\multicolumn{2}{c}{\textbf{Baby Products}} \\
\cmidrule(lr){2-3} \cmidrule(lr){4-5} \cmidrule(lr){6-7}
 & Recall@10 & NDCG@10 & Recall@10 & NDCG@10 & Recall@10 & NDCG@10 \\
\midrule
BERT  & 0.0214 & 0.0116 & 0.0157 & 0.0098 & 0.0122 & 0.0063 \\
Blair-Base  & \underline{0.0241} & 0.0128 & 0.0245 & 0.0123 & 0.0146 & 0.0131 \\
Blair-Large & 0.0247 & \underline{0.0134} & 0.0239 & 0.0119 & \underline{0.0148} & \underline{0.0075} \\
MiniLM-L6-v2         & 0.0231 & 0.0123 & 0.0218 & 0.0130 & 0.0130 & 0.0066 \\
EncodeRec-22M & 0.0232 & 0.0126 & \underline{0.0333} & \underline{0.0182} & 0.0141 & 0.0072 \\
EmbeddingGemma        & 0.0244 & 0.0129 & 0.0308 & 0.0165 & 0.0147 & 0.0075 \\
EncodeRec-300M & \textbf{0.0263*} & \textbf{0.0142*} & \textbf{0.0387*} & \textbf{0.0206*} & \textbf{0.0162*} & \textbf{0.0084*} \\
\bottomrule
\end{tabular}
}
\caption{UniSRec results across datasets. Best scores in \textbf{bold}, second best \underline{underlined}. The asterisk (*) indicates statistically significant improvements at $p < 0.05$ over the baselines.}

\label{tab:unsirec}
\end{table}

\begin{table}[t]
\centering
\renewcommand{\arraystretch}{1.6}
\resizebox{\columnwidth}{!}{%
\begin{tabular}{lcccccc}
\toprule
\multirow{2}{*}{\textbf{TIGER Model}} &
\multicolumn{2}{c}{\textbf{Sports}} &
\multicolumn{2}{c}{\textbf{Beauty}} &
\multicolumn{2}{c}{\textbf{Toys}} \\
\cmidrule(lr){2-3} \cmidrule(lr){4-5} \cmidrule(lr){6-7}
 & Recall@10 & NDCG@10 & Recall@10 & NDCG@10 & Recall@10 & NDCG@10 \\
\midrule
Sentence-T5      & 0.0382 & 0.0199 & 0.0601 & 0.0322 & 0.0578 & 0.0295 \\
Blair-Base       & 0.0270 & 0.0149 & 0.0546 & 0.0309 & 0.0583 & 0.0310 \\
Blair-Large      & 0.0312 & 0.0170 & 0.0599 & 0.0316 & 0.0546 & 0.0294 \\
MiniLM-L6-v2              & 0.0370 & 0.0198 & 0.0640 & 0.0350 & 0.0597 & 0.0312 \\
EncodeRec-22M    & \underline{0.0389} & \underline{0.0204} & \underline{0.0689} & \underline{0.0370} & \textbf{0.0663*} & \underline{0.0348} \\
EmbeddingGemma   & 0.0380 & 0.0203 & 0.0656 & 0.0353 & 0.0609 & 0.0331 \\
EncodeRec-300M   & \textbf{0.0395} & \textbf{0.0212} & \textbf{0.0694*} & \textbf{0.0373*} & \underline{0.0627} & \textbf{0.0362} \\
\bottomrule
\end{tabular}
}
\caption{TIGER results across datasets. Best scores in \textbf{bold}, second best \underline{underlined}.The asterisk (*) indicates statistically significant improvements at $p < 0.05$ over the baselines.}
\label{tab:tiger}
\end{table}

\section{Conclusion and Future Work}
We presented EncodeRec, a recommendation-oriented embedding backbone that departs from general-purpose PLM representations. By optimizing directly for recommendation signals, EncodeRec produces higher-resolution, domain-aware item embeddings. Across both sequential and generative paradigms, our results show consistent improvements, demonstrating the benefit of embeddings explicitly designed for recommendation tasks.

\noindent\textbf{Future Work.} Key next steps include (i) integrating richer user–item interaction signals during embedding training and (ii) extending the approach to multi-modal item features. These directions can further enhance the effectiveness and applicability of recommendation-centric embedding models.

\bibliographystyle{ACM-Reference-Format}
\bibliography{sample-base}

\end{document}